\documentclass{article}

    \PassOptionsToPackage{numbers, compress}{natbib}
\usepackage[preprint]{neurips_2025}

\usepackage[utf8]{inputenc} % allow utf-8 input
\usepackage[T1]{fontenc}    % use 8-bit T1 fonts
\usepackage{hyperref}       % hyperlinks
\usepackage{url}            % simple URL typesetting
\usepackage{booktabs}       % professional-quality tables
\usepackage{amsfonts}       % blackboard math symbols
\usepackage{nicefrac}       % compact symbols for 1/2, etc.
\usepackage{microtype}      % microtypography
\usepackage{xcolor}         % colors
\usepackage{graphicx}
\usepackage[most]{tcolorbox}
\usepackage{xcolor}
\usepackage{amsmath}
\usepackage{multirow}
\usepackage[capitalize]{cleveref}
\usepackage{wrapfig}
\usepackage[normalem]{ulem}
\useunder{\uline}{\ul}{}
\usepackage{amssymb}
\usepackage{caption}

\title{NOCL: Node-Oriented Conceptualization LLM for Graph Tasks without Message Passing}

\author{%
  Wei Li \\
  College of Computing and Data Science\\
  Nanyang Technological University \\
  \texttt{wei014@ntu.edu.sg} \\
  \And
  Mengcheng Lan \\
  College of Computing and Data Science\\
  Nanyang Technological University \\
  \texttt{lanm0002@e.ntu.edu.sg}
  \And
  Jiaxing Xu \\
  College of Computing and Data Science\\
  Nanyang Technological University \\
  \texttt{jiaxing003@e.ntu.edu.sg}
  \And
  Yiping Ke \\
  College of Computing and Data Science\\
  Nanyang Technological University \\
  \texttt{ypke@ntu.edu.sg}
}

\begin{document}

\maketitle

\begin{abstract}
Graphs are essential for modeling complex interactions across domains such as social networks, biology, and recommendation systems. 
Traditional Graph Neural Networks, particularly Message Passing Neural Networks (MPNNs), rely heavily on supervised learning, limiting their generalization and applicability in label-scarce scenarios. 
Recent self-supervised approaches still require labeled fine-tuning, limiting their effectiveness in zero-shot scenarios. 
Meanwhile, Large Language Models (LLMs) excel in natural language tasks but face significant challenges when applied to graphs, including preserving reasoning abilities, managing extensive token lengths from rich node attributes, and being limited to textual-attributed graphs (TAGs) and a single level task.
To overcome these limitations, we propose the Node-Oriented Conceptualization LLM (NOCL), a novel framework that leverages two core techniques: 
1) \textit{node description}, which converts heterogeneous node attributes into structured natural language, extending LLM from TAGs to non-TAGs;
2) \textit{node concept}, which encodes node descriptions into compact semantic embeddings using pretrained language models, significantly reducing token lengths by up to 93.9\% compared to directly using node descriptions.
Additionally, our NOCL employs \textit{graph representation descriptors} to unify graph tasks at various levels into a shared, language-based query format, paving a new direction for Graph Foundation Models. 
Experimental results validate NOCL's competitive supervised performance relative to traditional MPNNs and hybrid LLM-MPNN methods and demonstrate superior generalization in zero-shot settings.
  % The abstract paragraph should be indented \nicefrac{1}{2}~inch (3~picas) on
  % both the left- and right-hand margins. Use 10~point type, with a vertical
  % spacing (leading) of 11~points.  The word \textbf{Abstract} must be centered,
  % bold, and in point size 12. Two line spaces precede the abstract. The abstract
  % must be limited to one paragraph.
\end{abstract}

\section{Introduction}
Graphs, as versatile data structures, have become essential in modeling complex systems across diverse fields in natural and social sciences. Numerous real-world scenarios can be effectively represented as graphs, including social networks \citep{tang2009social}, brain networks \citep{xu2024contrasformer}, recommendation systems \citep{ying2018graph, ma2019learning}, protein interactions \citep{hamilton2017inductive}, fraud detection \citep{dou2020enhancing}, and traffic networks \citep{wu2020connecting, gao2020vectornet}. Message Passing Neural Networks (MPNNs) have emerged as useful tools for analyzing graph-structured data, primarily due to their ability to leverage structural dependencies and propagate information across nodes through message passing mechanisms.

However, standard MPNNs like GCN \citep{kipf2016semi} and GAT \citep{velivckovic2017graph} typically rely heavily on supervised learning, which limits their robustness and generalization across different datasets. Each new dataset often necessitates retraining the model entirely, hindering their practical deployment. Self-Supervised Learning (SSL) approaches, such as GCA \citep{zhu2021graph}, GraphMAE \citep{hou2022graphmae}, and S2GAE \citep{tan2023s2gae}, have emerged to enhance the generalization ability of MPNNs by pre-training models on unlabeled data through auxiliary tasks. Nevertheless, SSL methods still require fine-tuning with labeled data specific to downstream scenarios, limiting their applicability in contexts where high-quality labels are scarce or unavailable.

Recently, leveraging Large Language Models (LLMs) for graph-based applications has gained attention \citep{jin2024large}, due to their impressive generalization capabilities demonstrated in natural language processing (NLP) tasks. These models, such as GraphGPT \citep{tang2024graphgpt}, GIANT \citep{chien2021node}, and GLEM \citep{zhao2022learning}, primarily focus on textual-attributed graphs (TAGs) and node classification tasks. Other methods such as LLM-ICL \citep{guo2023indeed} and LLM4Mol \citep{qian2023can} integrate LLMs with molecule graph data by utilizing simplified molecular input line entry systems, but they neglect critical molecular node features like radical electrons and chirality, limiting their generalization. Moreover, a common challenge faced by these methods is the increased computational overhead due to longer token sequences, restricting practical applicability and scalability.

To effectively integrate LLMs with graph tasks, we identify three critical challenges:
\begin{itemize}
\item \textbf{C1.} Extending LLM's applicability from TAGs to non-TAGs.
\item \textbf{C2.} Maximizing the utilization of raw textual node features while maintaining manageable token lengths.
\item \textbf{C3.} Preserving LLM's reasoning and generalization capabilities across all graph tasks.
\end{itemize}

To address the aforementioned challenges, we propose Node-Oriented Conceptualization LLM (NOCL) with two key techniques: \textit{node description} and \textit{node concept}. The node description translates diverse node features into natural language paragraphs, thereby generalizing LLM's usage from TAG to non-TAG scenarios. Our node concept uses a pretrained language model (PLM) to encode node descriptions into compact embeddings, significantly reducing token lengths while preserving rich contextual information. Additionally, we introduce the graph representation descriptors to represent graph structures as textual sequences and reformulate all downstream graph tasks into human-readable queries. By transforming graph tasks into text-based comprehension problems, NOCL aligns naturally with the next-token prediction paradigm of LLMs, allowing them to directly generate task-specific outputs without relying on specialized architectural heads or rigid task-specific formats. This design not only supports standard classification tasks but also opens the door to open-ended reasoning over graphs, such as generating natural language explanations or answering free-form graph-based questions. While our current experiments focus on fixed-format tasks, this expressiveness represents a step toward more adaptive and general-purpose Graph Foundation Models (GFMs). Furthermore, by leveraging the inherent reasoning and generalization capabilities of LLMs, NOCL demonstrates strong zero-shot abilities—handling novel tasks or adapting to unseen domains without retraining. Architecturally, NOCL takes a deliberate step away from traditional MPNNs, which suffer from known limitations such as oversmoothing, locality bias, and difficulty in modeling long-range dependencies. By adopting an MPNN-free approach, NOCL offers a more flexible, scalable framework for reasoning over graph-structured data.

Our contributions can be summarized as follows:

\begin{itemize}
    \item We introduce a novel MPNN-free method, named NOCL. With our \textit{node description} and \textit{node concept} components, we incorporate rich semantic node features of both TAG and non-TAG into LLMs, while significantly reducing input token lengths, facilitating model tuning on commercial-grade hardware. %Besides, NOCL introduce a unified prompt-based approach that reformulates diverse graph tasks as language modeling problems, enabling seamless alignment of LLMs with multi-level graph understanding.
    \item Our approach aligns LLM with graph tasks at various levels and various types of graphs uniformly, paving a new way for comprehensive GFMs without MPNNs.
    \item Experimental results validate NOCL competitive supervised performance relative to traditional MPNNs and hybrid LLM-MPNN methods and demonstrate superior generalization in zero-shot settings.
\end{itemize}
\section{Preliminaries}
\textbf{Graph-structured data} represents information as entities (nodes) and the relationships (edges) between them. In this paper, we use $\mathcal{G} = (\mathcal{V}, \mathbf{X}, \mathbf{A})$ to denote a graph, where $\mathcal{V}=\{v_1, v_2, ..., v_n\}$ is the set of $n$ nodes. Each node $v_i \in \mathcal{V}$ is associated with a $d$-dimensional feature vector $x_i \in \mathbb{R}^d $ and $\mathbf{X} \in \mathbb{R}^{n\times d}$ denotes the node feature matrix. $\mathbf{A} \in \{0,1\}^{n \times n}$ represents the adjacency matrix, where $a_{ij}=1$ if there is an edge between nodes $v_i$ and $v_j$, and 0 otherwise. For node-level tasks, every node is associated with a label. For graph-level tasks, the graph is associated with
a label. For edge-level tasks, every two nodes are associated with a label to indicate whether an edge exists between these two nodes.

\section{Methodology}

\subsection{Motivation}

The pursuit of a universal GFM capable of reasoning across diverse graph types and solving a wide array of graph tasks is a crucial next step in the field \citep{mao2024position}. Inspired by the success of LLMs, recent efforts have explored their use in graph learning. However, building a truly general-purpose GFM remains an open challenge. As summarized in \cref{tab:GFM_Comparison}, existing LLM-based approaches often fall short by primarily focusing on TAGs and tackling individual tasks in isolation. Furthermore, many struggle with zero-shot generalization and rely on MPNNs, which have inherent limitations. To address these shortcomings and pave the way for a more versatile GFM, we propose a novel, MPNN-free model that supports a wide range of graph types and tasks within a unified framework. Our approach allows the LLM to generate task-specific outputs directly, without relying on specialized architectural heads. Beyond standard tasks, our model also opens new avenues for open-ended tasks, like reasoning and explanations—e.g., explaining why a node belongs to a particular class or which functional group contributes to a molecule’s specific properties.

\begin{figure}
  \centering
  \includegraphics[width=0.8\textwidth]{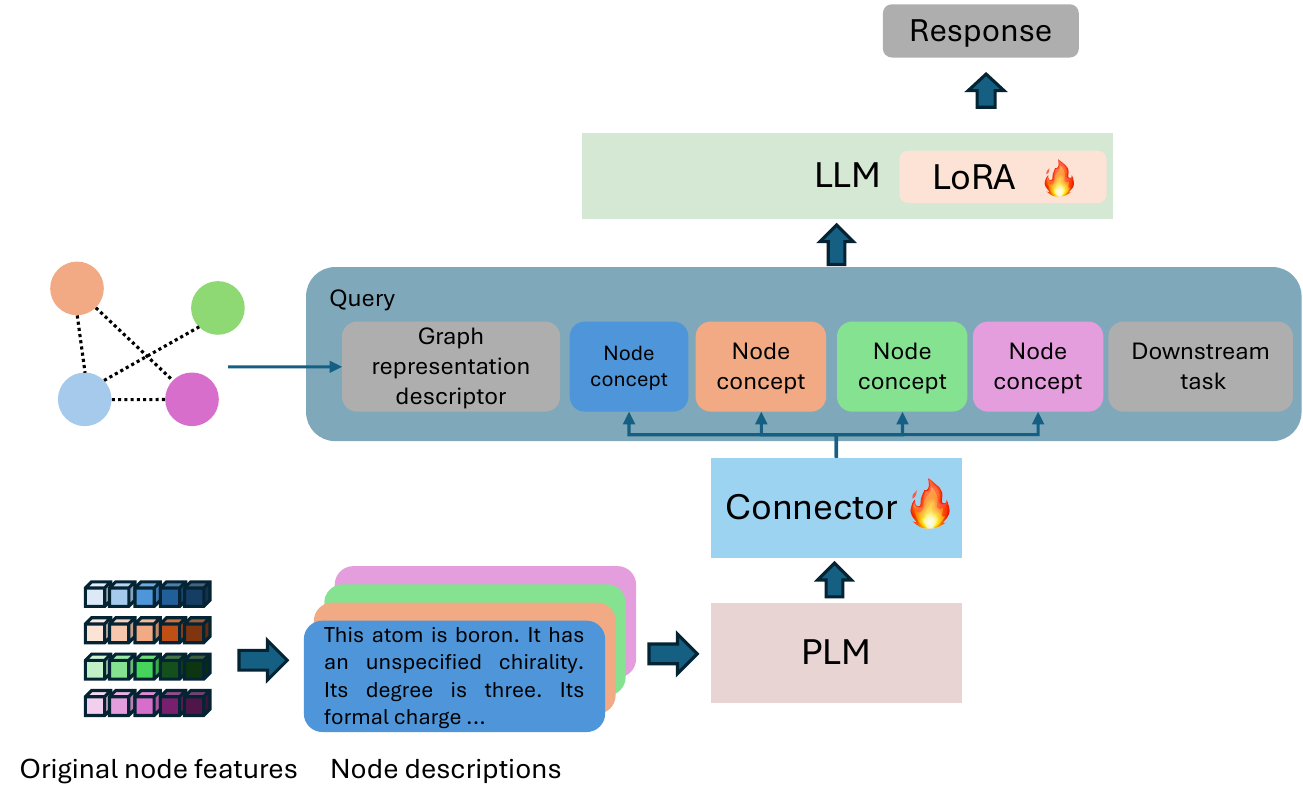}
  \caption{The overall framework of NOCL. The training process of NOCL consists of two stages. During the connector tuning stage, we freeze all parameters of the LLM and train only the connector module to align the node concept embeddings with the LLM input space. In the instruction tuning stage for downstream tasks, we freeze the connector and fine-tune the LLM using LoRA, enabling efficient adaptation to graph-related tasks with minimal additional parameters.}
  \label{fig:NOCL framework}
\end{figure}

The overall framework of our NOCL is illustrated in \cref{fig:NOCL framework}. We begin by converting original node features into node descriptions. These descriptions are then encoded into compact node concept embeddings using a PLM followed by a lightweight connector module. Next, we construct a prompt by integrating the graph representation descriptors, the encoded node concepts, and a textual description of the downstream task. This prompt is fed into the LLM, which generates the task-specific response directly through its next-token prediction capabilities.
\begin{table}[]
\centering
\caption{Comparison of capabilities of representative LLM-based graph learning models. Our NOCL offers the most comprehensive capabilities.}
\label{tab:GFM_Comparison}
\resizebox{\textwidth}{!}{%
\begin{tabular}{c|cc|c|ccc|c}
\toprule
 \multirow{2}{*}{Methods} &
  \multicolumn{2}{c|}{Graph Type} &
  \multirow{2}{*}{MPNN-Free} &
  \multicolumn{3}{c|}{Tasks} &
  \multirow{2}{*}{Open-end Potential} \\ \cline{2-3} \cline{5-7}
            & \multicolumn{1}{c}{TAG} & non-TAG &   & \multicolumn{1}{c}{node} & \multicolumn{1}{c}{link} & graph &   \\
\midrule
LLM-GNN \citep{chen2023label} &
  \multicolumn{1}{c}{\textbf{\checkmark}} &
  \textbf{} &
  \textbf{} &
  \multicolumn{1}{c}{\textbf{\checkmark}} &
  \multicolumn{1}{c}{\textbf{}} &
  \textbf{} &
  \textbf{} \\ %\hline
TAPE \citep{he2023harnessing}       & \multicolumn{1}{c}{\checkmark}   &         &   & \multicolumn{1}{c}{\checkmark}    & \multicolumn{1}{c}{}     &       &   \\ %\hline
InstructGLM \citep{ye2023language} & \multicolumn{1}{c}{\checkmark}   &         & \checkmark & \multicolumn{1}{c}{\checkmark}    & \multicolumn{1}{c}{\checkmark}    &       & \checkmark \\ %\hline
GraphGPT \citep{tang2024graphgpt}    & \multicolumn{1}{c}{\checkmark}   &         &   & \multicolumn{1}{c}{\checkmark}    & \multicolumn{1}{c}{\checkmark}    &       & \checkmark \\ %\hline
Mol-Instruction \citep{fang2023mol} &
  \multicolumn{1}{c}{\textbf{}} &
  \textbf{\checkmark} &
  \textbf{\checkmark} &
  \multicolumn{1}{c}{\textbf{}} &
  \multicolumn{1}{c}{\textbf{}} &
  \textbf{\checkmark} &
  \textbf{\checkmark} \\ %\hline
InstructMol \citep{cao2023instructmol} & \multicolumn{1}{c}{}    & \checkmark       &   & \multicolumn{1}{c}{}     & \multicolumn{1}{c}{}     & \checkmark     &   \\ %\hline
LLM4Mol \citep{qian2023can}     & \multicolumn{1}{c}{}    & \checkmark       & \checkmark & \multicolumn{1}{c}{}     & \multicolumn{1}{c}{}     & \checkmark     & \checkmark \\ %\hline
ReLM \citep{shi2023relm}        & \multicolumn{1}{c}{}    & \checkmark       &   & \multicolumn{1}{c}{}     & \multicolumn{1}{c}{}     & \checkmark     & \checkmark \\ %\hline
MolCA \citep{liu2023molca}      & \multicolumn{1}{c}{}    & \checkmark       &   & \multicolumn{1}{c}{}     & \multicolumn{1}{c}{}     & \checkmark     & \checkmark \\ %\hline
\midrule
NOCL (ours)       & \multicolumn{1}{c}{\checkmark}   & \checkmark       & \checkmark & \multicolumn{1}{c}{\checkmark}    & \multicolumn{1}{c}{\checkmark}    & \checkmark     & \checkmark \\
\bottomrule
\end{tabular}%
}
\end{table}
% \begin{table}[]
% \centering
% \caption{Comparison between NOCL and other LLM for graph methods}
% \label{tab:GFM_Comparison}
% \resizebox{\textwidth}{!}{%
% \begin{tabular}{@{}|c|cc|c|ccc|c|c|@{}}
% \toprule
%  & \multicolumn{2}{c|}{Graph Type} & \multirow{2}{*}{MPNN-Free} & \multicolumn{3}{c|}{Tasks} & \multirow{2}{*}{Generalization} & \multirow{2}{*}{Open-end Potential} \\ \cmidrule(r){1-3} \cmidrule(lr){5-7}
%             & \multicolumn{1}{c|}{TAG} & non-TAG &   & \multicolumn{1}{c|}{node} & \multicolumn{1}{c|}{link} & graph &   &   \\ \midrule
% TAPE        & \multicolumn{1}{c|}{\checkmark}   &         &   & \multicolumn{1}{c|}{\checkmark}    & \multicolumn{1}{c|}{}     &       &   &   \\ \midrule
% InstructGLM & \multicolumn{1}{c|}{\checkmark}   &         & \checkmark & \multicolumn{1}{c|}{\checkmark}    & \multicolumn{1}{c|}{\checkmark}    &       & \checkmark & \checkmark \\ \midrule
% GraphGPT    & \multicolumn{1}{c|}{\checkmark}   &         &   & \multicolumn{1}{c|}{\checkmark}    & \multicolumn{1}{c|}{\checkmark}    &       & \checkmark & \checkmark \\ \midrule
% InstructMol & \multicolumn{1}{c|}{}    & \checkmark       &   & \multicolumn{1}{c|}{}     & \multicolumn{1}{c|}{}     & \checkmark     &   &   \\ \midrule
% NOCL        & \multicolumn{1}{c|}{\checkmark}   & \checkmark       & \checkmark & \multicolumn{1}{c|}{\checkmark}    & \multicolumn{1}{c|}{\checkmark}    & \checkmark     & \checkmark & \checkmark \\ \bottomrule
% \end{tabular}
% }
% \end{table}

\subsection{Node Description}
\begin{figure}
  \centering
  \includegraphics[width=1\textwidth]{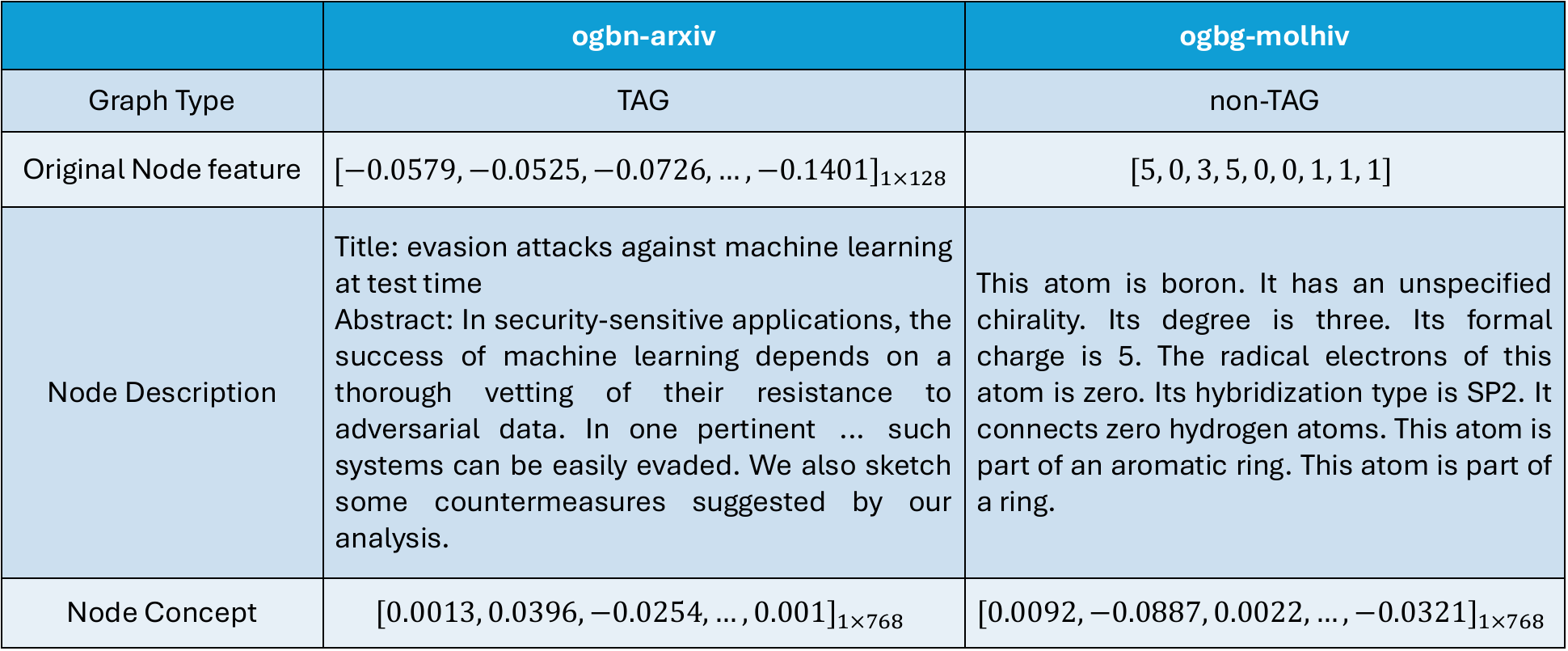}
  \caption{Node Concept Examples.}
  \label{fig:node_concept_examples}
\end{figure}

The node description is a natural language representation of a node. As illustrated in \cref{fig:node_concept_examples}, TAGs directly utilize original texts as node descriptions. In contrast, non-TAGs, such as molecular graphs, require generating descriptive language to characterize node features. For instance, in the ogbg-molhiv dataset, each node corresponds to an atom and is associated with an 8-dimensional feature vector, capturing attributes such as atomic number, chirality, formal charge, and others. We convert each atom's original numerical features into natural language using the following template:
\begin{quote}
    This atom is [atomic name]. It has a [chirality type]. Its formal charge is [formal charge number]. The radical electrons of this atom is [number of radical electrons]...% Its hybridization type is [hybridization type]... % It connects [number of hydrogen atoms] hydrogen atoms...
\end{quote}
We provides more details and examples about generating node descriptions in \cref{app:node_descriptions}.

\subsection{Node Concept}

\textbf{Definition of node concept.} Our \textit{node concept} refers to a fixed-size embedding vector derived from encoding a node description using a PLM, such as BERT \citep{devlin2019bert}. 

A straightforward approach to leverage node descriptions would be directly inputting them into LLMs. However, this approach faces significant limitations due to lengthy token sequences, especially prevalent in TAGs. For instance, the average node description length for the ogbn-arxiv dataset \citep{hu2020ogb} is 221 tokens, requiring approximately 1 second per node for classification on an RTX 4090 GPU. Moreover, when handling graphs with numerous nodes, the aggregated token lengths quickly exceed the maximum context window of LLMs (e.g., 4096 tokens in Llama 2 \citep{touvron2023llama}).

To address these challenges, we introduce the node concept embedding, which effectively reduces node descriptions into compact embedding vectors via PLMs. Our proposed approach facilitates seamless integration of graph-structured data into multimodal LLMs (MLLMs) by:
\begin{itemize}
\item Requiring no architectural modifications, ensuring easy adoption and scalability.
\item Significantly reducing token lengths, thus simplifying optimization and boosting efficiency.
\end{itemize}

\begin{figure}[h]
    \centering
    % \vspace{-5pt}
    \includegraphics[width=1\textwidth]{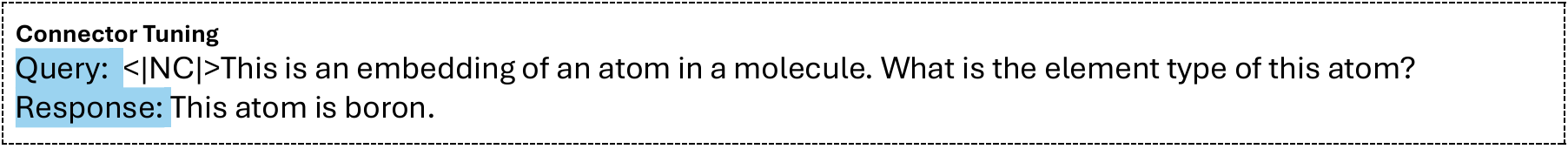}
    \caption{Connector Tuning Example.}
    \label{fig:connector_tuning}
    % \vspace{-20pt}
\end{figure}

\textbf{Node concept connector tuning.} The primary goal of node concept connector is to effectively leverage the capabilities of both the PLM and pretraind LLM. For each node description, we generate multiple single-turn conversations data, 
as one example illustrated as \cref{fig:connector_tuning}.
%as one example illustrated in \cref{fig:connector_tuning}. 
 By treating all answers as the assistant's response and training the connector, the connector could align the embedding spaces of PLM and LLM together. When we train the connector, the LLM is frozen. More details and examples of connector tuning are provided in \cref{app:connector_tuning}.

\subsection{Graph Task Instruction Tuning}

\textbf{Downstream tasks reformulation} Apart from varying embedding spaces across datasets, another challenge for MPNNs is their inability to uniformly integrate graph tasks at multiple levels within a single framework. For instance, node classification tasks typically involve directly training a classifier on node embeddings. However, graph-level classification tasks necessitate additional readout functions, such as sum pooling or mean pooling \citep{atwood2016diffusion, simonovsky2017dynamic}, to aggregate node-level information into a graph-level representation. This discrepancy constrains the performance of pre-trained models and may lead to negative transfer \citep{jin2020self}. To overcome these limitations and achieve a unified framework for all graph tasks, we adopt the prompt-learning approach proposed by \citet{sun2023all}. Specifically, we reformulate node-level and edge-level tasks into equivalent graph-level tasks by constructing induced graphs centered around nodes or edges. Consequently, all tasks are effectively transformed into graph-level comprehension tasks for LLMs:1) \textbf{Node-level}: Determining the category of a node within the graph. 2) \textbf{Edge-level}: Predicting whether two nodes should be connected within the graph. 3) \textbf{Graph-level}: Classifying the category of the entire graph. This unified reformulation leverages the next-token prediction framework intrinsic to MLLMs, exploiting their powerful text-generation capabilities for graph understanding.

\textbf{Graph representation descriptors.} We represent a graph as a sequential language structure that first generates all nodes along with their corresponding node concepts, followed by the edges. Specifically, a graph $\mathcal{G}$ with $n$ nodes and $m$ edges can be described as:
$$
\text{<|BON|>}\ \underbrace{\text{<|NC|>}_1\ 1\ \cdots\ \text{<|NC|>}_n\ n}_{2n }\ \text{<|EON|>}\ \text{<|BOE|>}
\ \underbrace{\text{<|EDGE|>}\ 1\ \, k \cdots \ \text{<|EDGE|>}\ i\ \, j}_{3m}\ \text{<|EOE|>}
$$
where $i,j, k \in \{1,2,\dots n\}$, $a_{1k} = 1$ and $a_{ij} = 1$.
Special tokens introduced include $\text{<|BON|>}$ (beginning of nodes), $\text{<|NC|>}$ (node concept placeholder), $\text{<|EON|>}$ (end of nodes), $\text{<|BOE|>}$ (beginning of edges), $\text{<|EDGE|>}$ (individual edge tuples), and $\text{<|EOE|>}$  (end of edges). The overall token length of  graph representation descriptors for $\mathcal{G}$ is $4+2n+3n$. Given that multiple sequence representations can arise from varying node permutations and edge generation orders, we standardize the sequence by placing the target node as the first node for node-level tasks. For edge-level tasks, we similarly position one target node $v_i$ first and generate all nodes within its induced graph $\mathcal{G}_{v_i}$. We subsequently generate the second target node $v_j$ along with its induced graph $\mathcal{G}_{v_j}$. Edges are generated randomly within graph descriptors, except for edge-level tasks, where edges from $\mathcal{G}_{v_i}$ precede edges from $\mathcal{G}_{v_j}$.

\textbf{Instruction tuning of downstream tasks.} Leveraging the proposed graph descriptors, we construct graph instruction datasets from existing structured datasets. Each <graph, task> pair is formatted into a query-response template:
\begin{quote}
    \textbf{Query:}
    % \begin{verbatim}
        This is a graph: <Graph Descriptors>. <Downstream Task Query>
    % \end{verbatim}
    \\
    \textbf{Response:}
    % \begin{verbatim}
        <Corresponding Text Label>
    % \end{verbatim}
\end{quote}

% \textbf{Query:} \texttt{This is a graph: <Graph Descriptors>.<Downstream Task Query>}

% \textbf{Response:} \texttt{<Corresponding Text Label>}

\begin{figure}
  \centering
  \includegraphics[width=1\textwidth]{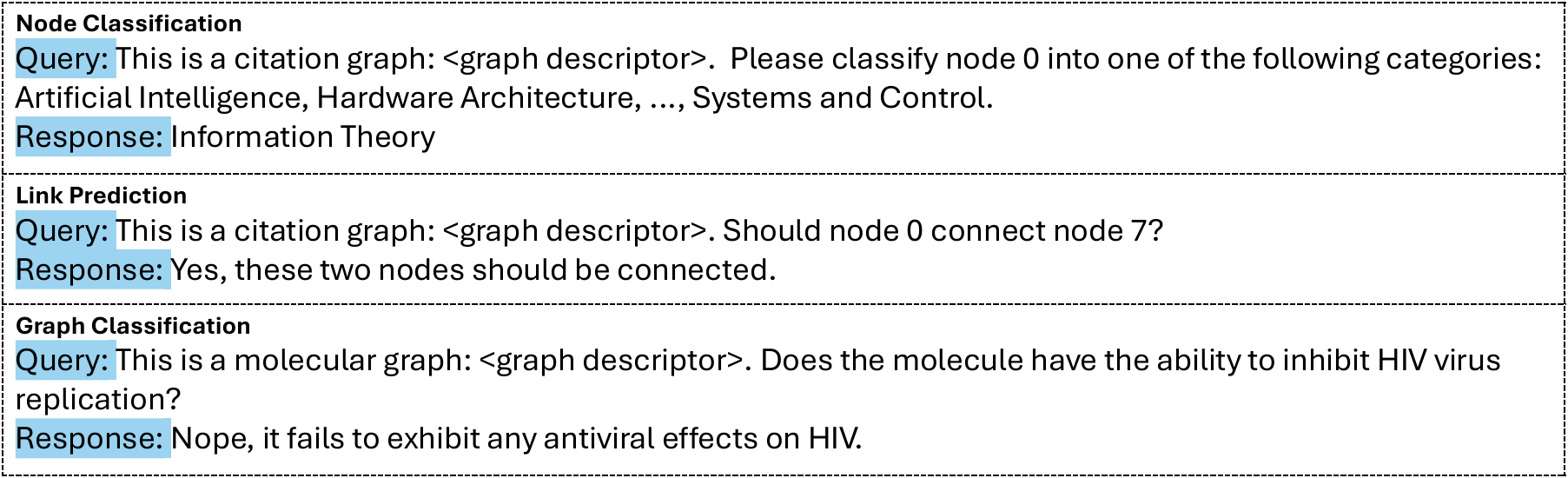}
  \caption{Graph Instruct Tuning Examples.}
  \label{fig:QA_examples}
\end{figure}

\Cref{fig:QA_examples} illustrates examples of queries and responses for various graph tasks. By utilizing pure textual responses, our NOCL approach enables seamless integration with existing MLLMs without architectural changes. We employ LoRA \citep{hu2022lora} to fine-tune LLMs on graph-structured instruction datasets, optimizing the original autoregressive training objective $\mathcal{L}_{txt}$ inherent to LLMs. To enhance the LLM's understanding of graph representation descriptors, we also incorporate node counting and edge checking problems as part of our downstream tasks. More examples of instruction tuning are provided in \cref{app:graph_tasks}.

% \textbf{Graph-level task} \cite{han2025rethinking}
% placeholder
\section{Experiments}
\subsection{Experimental Settings}
\textbf{Model architectures.} For our experiments, we employ the Llama-3.2-3B-Instruct and Llama-3.2-1B-Instruct models as our base LLM and utilize the \texttt{all-mpnet-base-v2} model from Sentence-BERT (S-BERT) \citep{reimers2019sentence} as the encoder for node descriptions. A simple linear layer serves as the connector bridging the node concept embeddings and the LLM.

\textbf{Datasets.} We evaluate our approach across five datasets: ogbn-arxiv \citep{hu2020ogb}, PubMed \citep{sen2008collective}, Cora \citep{mccallum2000automating}, MUTAG \citep{debnath1991structure} and ogbg-molhiv \citep{hu2020ogb}. The detail statistics and prompts of these datasets are provided in \cref{app:datasets}.  These datasets are categorized according to their respective tasks as follows: \textbf{Node level:} ogbn-arxiv, PubMed, Cora; \textbf{Link level:} PubMed, Cora; \textbf{Graph level:} MUTAG, ogbn-molhiv. We utilize all nodes from these datasets to generate the corresponding node descriptions and node concepts, which are then employed in training the connector.
% \textbf{Evaluation Protocols.} 
In our experiments, we simultaneously train models on tasks across multiple levels. For node-level tasks, we use only the training set of ogbn-arxiv. For link-level tasks, we adopt the evaluation protocol established by \citep{kipf2016variational, pan2018adversarially},  partitioning the links in the Cora dataset into 85\% training, 5\% validation, and 10\% testing subsets. An equal number of non-existing (negative) links are randomly sampled and added to each subset. For graph-level tasks, we use 80\% of the MUTAG dataset and the training set of ogbg-molhiv for training purposes. Datasets not explicitly mentioned are reserved exclusively for zero-shot evaluation. We carefully ensure that there is no data leakage across the datasets utilized for different multi-level tasks. 

\textbf{Metrics.} To evaluate our model's performance, we utilize two commonly used metrics: Accuracy for node classification and ROC\_AUC for link prediction and graph classification. 

\textbf{Baselines.} In our performance comparison, we consider various state-of-the-art methods for comprehensive evaluation. (i) MPNNs, including Graph-SAGE \citep{hamilton2017inductive}, GCN \citep{kipf2016semi}, GAT \citep{velivckovic2017graph}, RevGNN \citep{li2021training}, GKD \citep{yang2022geometric} and GLNN \citep{zhang2021graph}. (ii) open LLMs, such as Baichuan-7B \citep{baichuan2023b}, vicunas \citep{chiang2023vicuna}, Galactica \citep{taylor2022galactica} and Llama-3.2s \citep{meta2024llama3}, to directly understanding TAGs. (iii) LLM-MPNNs, including: GraphGPT \citep{tang2024graphgpt}, InstructGLM \citep{ye2023language} and InstructMol \citep{cao2023instructmol}.

\textbf{Implementation details.} 
All of our models are trained using 4 NVIDIA RTX 4090 GPUs (24GB each). During the tuning phase of the node concept connector, we set the batch size to 1 per device and use a learning rate of 3e-1. For instruction tuning on downstream tasks, we use a batch size of 2 per device, a learning rate of 3e-3, and apply LoRA with a rank of 16 for fine-tuning. For inducing subgraphs in node- and link-level tasks, we use only 1-hop neighbors and ensure that the total number of nodes in the subgraph does not exceed 11. If a target node has more than 10 neighbors, we randomly sample 10 to construct the induced subgraph. To adapt LLM outputs for the ROC\_AUC evaluation metric, we flatten the last hidden state corresponding to the first output token of the LLM and apply a linear projection to produce the final numeric prediction. The more details, such as other hyperparameters and training hours, are provided in \cref{app:experiments}.

\begin{table}[t]
\centering
\footnotesize
\setlength{\tabcolsep}{10pt}
\caption{Performance comparison of various methods on node classification under both supervised and zero-shot settings. * indicates that results are obtained from the GraphGPT papers. 
%- indicates there is no zero-shot result of original methods. 
The best results are in \textbf{bold}, and the second best are {\ul underlined}.}
\label{tab:node_classification}
\begin{tabular}{c|c|c|c|c}
\toprule
\multirow{2}{*}{Methods}          &                            & ogbn-arxiv      & PubMed          & Cora            \\ \cline{2-5} 
                           & Training method            & supervised      & zero-shot       & zero-shot       \\
\midrule
\multirow{6}{*}{MPNNs}     & GraphSAGE*                 & 0.5480          & 0.3950          & 0.0328          \\ %\cline{2-5} 
                           & GCN*                       & 0.5267          & 0.3940          & 0.0214          \\ %\cline{2-5} 
                           & GAT*                       & 0.5332          & 0.3940          & 0.0167          \\ %\cline{2-5} 
                           & RevGNN*                    & 0.5474          & 0.4440          & 0.0272          \\ %\cline{2-5} 
                           & GKD*                       & 0.5570          & 0.3645          & 0.0470          \\ %\cline{2-5} 
                           & GLNN*                      & 0.6088          & 0.4298          & 0.0267          \\
\midrule
\multirow{4}{*}{LLMs}      & Baichuan-7B*               & 0.0946          & 0.4642          & 0.0405          \\ %\cline{2-5} 
                           & vicuna-7B-v1.5*            & 0.4962          & {\ul 0.6351}    & 0.1489          \\ %\cline{2-5} 
                           & Llama-3.2-1B-Instruct      & 0.0549          & 0.0000          & 0.1809          \\ %\cline{2-5} 
                           & Llama-3.2-3B-Instruct      & 0.5141          & 0.0002          & {\ul 0.5096}    \\
\midrule
\multirow{2}{*}{MPNN-LLMs} & GraphGPT-vicuna-7B-1.5     & 0.6476          & \textbf{0.7011} & 0.1813          \\ %\cline{2-5} 
                           & InstructGLM-Flan-T5-large  & {\ul 0.7467}    & -               & -               \\
\midrule
\multirow{2}{*}{NOCL (ours)}      & NOCL-Llama-3.2-1B-Instruct & 0.7440          & 0.2820          & 0.3231          \\ %\cline{2-5} 
                           & NOCL-Llama-3.2-3B-Instruct & \textbf{0.7478} & 0.4764          & \textbf{0.5583} \\
\bottomrule
\end{tabular}
\end{table}

\begin{table}[t]
\centering
\footnotesize
\setlength{\tabcolsep}{10pt}
\caption{Performance on link and graph tasks. * indicates that results are obtained under the zero-shot setting.}
\label{tab:link_graph}
% \resizebox{\textwidth}{!}{%
\begin{tabular}{c|cc|cc}
\toprule
\multirow{3}{*}{Methods} & \multicolumn{1}{c|}{Cora}       & PubMed          & \multicolumn{1}{c|}{ogbg-molhiv} & MUTAG      \\ \cline{2-5} 
                  & \multicolumn{2}{c|}{Link}                         & \multicolumn{2}{c}{Graph}                     \\ \cline{2-5} 
                  & \multicolumn{1}{c|}{supervised} & supervised      & \multicolumn{1}{c|}{supervised}  & supervised \\
\midrule
GraphSAGE         & \multicolumn{1}{c|}{0.6789}     & 0.6439          & \multicolumn{1}{c|}{0.7171}      & 0.6310     \\
GCN               & \multicolumn{1}{c|}{0.6546}     & {\ul 0.6853}    & \multicolumn{1}{c|}{0.7256}      & 0.5357     \\ 
GAT               & \multicolumn{1}{c|}{0.6591}     & \textbf{0.7064} & \multicolumn{1}{c|}{0.7371}      & 0.5357     \\ 
\midrule
NOCL-Llama-3.2-1B-Instruct & \multicolumn{1}{c|}{{\ul 0.8842}}    & 0.6057* & \multicolumn{1}{c|}{{\ul 0.7476}} & \textbf{0.7262} \\ 
NOCL-Llama-3.2-3B-Instruct & \multicolumn{1}{c|}{\textbf{0.8965}} & 0.6351* & \multicolumn{1}{c|}{0.7576}       & {\ul 0.7023}    \\ \bottomrule
\end{tabular}
\end{table}

\subsection{Overall Performance}
\textbf{Node classification.} We evaluate our approach on node classification tasks under both supervised and zero-shot settings. Results are presented in \cref{tab:node_classification}. In the supervised setting, our NOCL outperforms state-of-the-art baselines. In the zero-shot setting, our proposed method achieves the best performance on the Cora dataset and ranks third on PubMed. The relatively lower performance on PubMed is likely due to the limitations of the base LLM, as both base LLMs demonstrate poor classification performance on this dataset.  Nevertheless, through the integration of node concepts, graph descriptor prompts, and downstream task instruction tuning, we significantly enhance the generalization capability of our base LLMs on all datasets, including PubMed.

\textbf{Link prediction. } As shown in \cref{tab:link_graph},  the proposed NOCL surpasses all MPNN baselines under the supervised setting. Remarkably, under the zero-shot setting, NOCL achieves a comparable performance to supervised GraphSAGE.
Besides, MPNNs employ cosine similarity for determining edge existence, leveraging their intrinsic smoothing properties without the need for explicit link prediction. 
When we modify MPNNs to concatenate the outputs of target node pairs and train a classifier for link prediction, their performance drops to be around 0.55 in ROC\_AUC even in supervised settings, underscoring the competitiveness of our model.

\textbf{Graph classification.} As shown in \cref{tab:link_graph}, NOCL also consistently outperforms MPNNs across all evaluated datasets. Notably, on the MUTAG dataset, our approach exceeds the best-performing MPNN by a margin of 0.09 in ROC\_AUC, demonstrating superior performance in graph-level tasks. 
Additionally, we provide a comparison for NOCL with other LLM-based methods on ogbg-molhiv dataset. As shown in \cref{tab:molhiv_only}, our method outperforms all other methods.

\subsection{Contribution of Graph Tasks at Various Levels}

We conduct an study to investigate the contributions of different graph tasks for our proposed framework. The results are reported in \cref{tab:ablation}. All variants are based on Llama-3.2-1B-Instruct and the task is node classification. We denote: ``SD'': fine-tuned on a single dataset (ogbn-arxiv) only; ``GU'': fine-tuned with additional graph structure understanding tasks, including node counting and edge checking; ``Mix'': fine-tuned with a combination of graph tasks across different levels (node, edge, graph).
The following observations emerge: 1) Incorporating graph structure understanding tasks improves the model’s generalization ability. While the SD+GU variant performs slightly worse than SD in the supervised setting, it demonstrates significantly better zero-shot performance—indicating that structural tasks enhance the LLM’s ability to generalize to unseen data. 2) Instruction mixing across task levels yields further improvements. The Mix variant, trained on a blend of node-level, edge-level, and graph-level tasks retains strong capability on single task of node classification. This suggests that NOCL benefits from multi-task instruction tuning without compromising single-task effectiveness.

\begin{figure}[t]
\begin{minipage}{0.5\textwidth}
\centering
\footnotesize
\captionof{table}{Performance on graph classification on ogbg-molhiv of LLM-based methods.}
        \label{tab:molhiv_only}
        \tabcolsep8pt
        \begin{tabular}{c|c}
        \toprule
                        Methods    & ogbg-molhiv     \\
        \midrule
        Galactica-6.7B             & 0.7220          \\ 
        Vicuna-v1.3-7B             & 0.5810          \\ 
        InstructMol-G              & 0.7400          \\  
        InstructMol-GS             & 0.6890          \\
        \midrule
        NOCL-Llama-3.2-1B-Instruct & {\ul 0.7476}    \\
        NOCL-Llama-3.2-3B-Instruct & \textbf{0.7576} \\
        \bottomrule
        \end{tabular}
\end{minipage}
\hfill
\begin{minipage}{0.45\textwidth}
\centering
\footnotesize
\captionof{table}{Contribution study on node classification under both supervised and zero-shot settings.}
\label{tab:ablation}
\tabcolsep6pt
\begin{tabular}{c|c|c}
\toprule
 Variants & \begin{tabular}[c]{@{}c@{}}ogbn-arxiv\\  supervised\end{tabular} & \begin{tabular}[c]{@{}c@{}}PubMed\\  zero-shot\end{tabular} \\
 \midrule
SD            & 0.7424          & 0.2462          \\
SD+GU         & 0.7291          & \textbf{0.2881} \\
\midrule
NOCL (GU+Mix) & \textbf{0.7440} & 0.2820          \\
\bottomrule
\end{tabular}
\end{minipage}

\end{figure}

\subsection{Model Efficiency Study}

Our proposed node concept significantly reduces input sequence token length during inference and training. As shown in \cref{tab:token-size count}, it decreases the average sequence length from thousands to hundreds of tokens, achieving a reduction ratio of at least 81.7\%. This enables highly efficient memory and time usage. For example, one data point training with a NOCL-Llama-3.2-1B-Instruct on the PubMed dataset reduces memory requirements from approximately 12.10 GB to 2.63 GB. As a result, both training and inference become feasible on commercial-grade GPUs such as the RTX 4090 (24 GB), enabling tasks, such as training on PubMed and ogbg-molhiv, that were previously infeasible due to memory constraints. Moreover, the shorter input sequences substantially accelerate training. For instance, when fine-tuning NOCL-Llama-3.2-1B-Instruct on the ogbn-arxiv dataset for one epoch, training time is reduced from 53.5 minutes (using full node descriptions) to just 8.7 minutes, achieving a 6.15× speedup.

\begin{table}[h]
\centering
\footnotesize
\caption{The statistics of average input token lengths, one-epoch training times, and GPU memory usage for training one random sampled data point across different configurations. The ``OOM'' indicates that the model ran out of memory during training, even with a batch size of 1. The base LLM model is \texttt{Llama-3.2-1B-Instruct}.}
\label{tab:token-size count}
\resizebox{\textwidth}{!}{%
\begin{tabular}{@{}c|c|c|cc|cc|cc@{}}
\toprule
Dataset &
  Task Level &
  \begin{tabular}[c]{@{}c@{}} Node Feature \\ (tokens) \end{tabular} &
  \multicolumn{2}{c|}{\begin{tabular}[c]{@{}c@{}}Question Sequence \\(tokens) \end{tabular}} &
  \multicolumn{2}{c|}{\begin{tabular}[c]{@{}c@{}}Training Time \\ (min)\end{tabular}} &
  \multicolumn{2}{c}{\begin{tabular}[c]{@{}c@{}}GPU Memory \\ (GB)\end{tabular}} \\ \midrule
ogbn-arxiv &
  \multirow{2}{*}{node} &
  221 &
  \multicolumn{1}{c|}{1891} &
  \multirow{2}{*}{83.77\%} &
  \multicolumn{1}{c|}{53.5} &
  \multirow{2}{*}{6.15x} &
  \multicolumn{1}{c|}{4.92} &
  \multirow{2}{*}{38.62\%} \\ \cmidrule(lr){3-4} \cmidrule(lr){6-6} \cmidrule(lr){8-8}
+ node concept &
   &
  1 &
  \multicolumn{1}{c|}{307} &
   &
  \multicolumn{1}{c|}{8.7} &
   &
  \multicolumn{1}{c|}{3.02} &
   \\ \midrule
PubMed &
  \multirow{2}{*}{node} &
  370 &
  \multicolumn{1}{c|}{1731} &
  \multirow{2}{*}{90.93\%} &
  \multicolumn{1}{c|}{OOM} &
  \multirow{2}{*}{-} &
  \multicolumn{1}{c|}{7.64} &
  \multirow{2}{*}{63.31\%} \\ \cmidrule(lr){3-4} \cmidrule(lr){6-6} \cmidrule(lr){8-8}
+ node concept &
   &
  1 &
  \multicolumn{1}{c|}{157} &
   &
  \multicolumn{1}{c|}{0.68} &
   &
  \multicolumn{1}{c|}{2.65} &
   \\ \midrule
PubMed &
  \multirow{2}{*}{link} &
  370 &
  \multicolumn{1}{c|}{3633} &
  \multirow{2}{*}{93.86\%} &
  \multicolumn{1}{c|}{OOM} &
  \multirow{2}{*}{-} &
  \multicolumn{1}{c|}{12.10} &
  \multirow{2}{*}{78.26\%} \\ \cmidrule(lr){3-4} \cmidrule(lr){6-6} \cmidrule(lr){8-8}
+ node concept &
   &
  1 &
  \multicolumn{1}{c|}{223} &
   &
  \multicolumn{1}{c|}{4.01} &
   &
  \multicolumn{1}{c|}{2.63} &
   \\ \midrule
ogbg-molhiv &
  \multirow{2}{*}{graph} &
  53 &
  \multicolumn{1}{c|}{1639} &
  \multirow{2}{*}{81.70\%} &
  \multicolumn{1}{c|}{OOM} &
  \multirow{2}{*}{-} &
  \multicolumn{1}{c|}{6.72} &
  \multirow{2}{*}{59.23\%} \\ \cmidrule(lr){3-4} \cmidrule(lr){6-6} \cmidrule(lr){8-8}
+ node concept &
   &
  1 &
  \multicolumn{1}{c|}{300} &
   &
  \multicolumn{1}{c|}{2.05} &
   &
  \multicolumn{1}{c|}{2.74} &
   \\ \bottomrule
\end{tabular}%
}
\end{table}

\section{Related Work}
\textbf{Multimodal Large Language Models.} MLLMs are typically developed by enhancing LLMs with visual perception modules, which can generate coherent textual conversations grounded in multimodal inputs. For instance, Flamingo \citep{alayrac2022flamingo} introduces
the Perceiver Resampler, which connects a pre-trained vision encoder with LLMs for effective few-shot learning.  BLIP-2 \citep{li2023blip} bridges the modality gap using a lightweight Querying Transformer (Q-Former), demonstrating enhanced performance on zero-shot vision-to-language tasks. The LLaVA seires \citep{liu2023visual,liu2024improved} employs a linear layer or MLP as a modality connector,
trained on multimodal language-image instruction-following data generated with GPT-4, showcasing notable capabilities in multimodal chat interactions. They demonstrate impressive capabilities
in multimodal chat interactions. In this work, we followed the LLaVA \citep{liu2023visual} structure and extend MLLM modality from images to graphs.

\textbf{LLM for Graphs.} Recently, there has been an increasing interest in extending LLMs for graph-based applications. Depending on the role of LLMs and their interaction with MPNNs,  \citet{jin2024large} have classified LLM for graphs into three categories: 1) LLM as a predictor. LLMs directly work as the final predictor for graph tasks. For instance, InstrcutGLM \citep{ye2023language} directly converts graph structures into natural languages and apply LLMs to node classification and link prediction. 2) LLM as an encoder. LLMs extract textual features to serve as initial node feature vectors for MPNNs, which then generate node/edge representations and make predictions. These methods typically adopt an LLM-GNN cascaded architecture to obtain the final representation, such as TAPE \citep{he2023harnessing}. 3) LLM as an aligner. These methods contain an LLM component for text encoding and a GNN component for structure encoding. These two components are served equally and trained iteratively or in parallel. LLMs and MPNNs can mutually enhance each other since the LLMs can provide textual signals to MPNNs, while the MPNNs can deliver structural information to LLMs. For example, GLEM \citep{zhao2022learning} formulates the iterative training process into a pseudo-likelihood variational framework, where the E-step is to optimize the LLM and the M-step is to train the GNN. Our work could be considered as using LLM as a predictor.

\section{Limitation} \label{limitation}

Our approach presents several potential limitations. First, the performance of NOCL is significantly influenced by the choice of the PLM. While \texttt{all-mpnet-base-v2} is a general-purpose model, it may struggle with underrepresented or domain-specific language expressions. Second, although our node concept strategy enables the extension of LLM applications from TAGs to non-TAGs, generating high-quality node descriptions for non-TAGs often requires expert knowledge or auxiliary tools. Third, due to computational constraints, our experiments are conducted on only five datasets and only consider up to 10 nodes in the 1-hop induced graph. It remains an open question whether the model can maintain strong generalization performance on a broader range of graph-structured data. We leave this investigation for future work.

\section{Conclusion}
In this work, we introduced a novel LLM-centric framework that reimagines graph representation and reasoning without relying on traditional MPNN architectures. By formulating the concepts of node description and node concept, we enable LLMs to effectively operate across both TAG and non-TAG graph domains. Our design addresses three critical challenges—preserving reasoning capabilities, controlling token overhead, and extending generalizability—through a unified text-based formulation of graph tasks. This not only broadens the applicability of LLMs in graph learning but also opens a new direction toward scalable, zero-shot Graph Foundation Models (GFMs). Experimental results demonstrate that our approach achieves strong performance in both supervised and zero-shot settings, while maintaining resource efficiency. Future work will explore deeper integration of temporal and dynamic graph contexts and further optimize instruction-tuning strategies to support increasingly complex graph reasoning scenarios.

\newpage

\bibliographystyle{ACM-Reference-Format}
\bibliography{neurips_2025}

\newpage
\section*{NeurIPS Paper Checklist}

\begin{enumerate}

\item {\bf Claims}
    \item[] Question: Do the main claims made in the abstract and introduction accurately reflect the paper's contributions and scope?
    \item[] Answer: \answerYes{} %\answerTODO{} % Replace by \answerYes{}, \answerNo{}, or \answerNA{}.
    \item[] Justification: We made the claim that NOCL could deal with graph tasks with varying levels and multi-type of graphs.  %\answerNA{} %\justificationTODO{}

\item {\bf Limitations}
    \item[] Question: Does the paper discuss the limitations of the work performed by the authors?
    \item[] Answer: \answerYes{} %\answerTODO{} % Replace by \answerYes{}, \answerNo{}, or \answerNA{}.
    \item[] Justification: In \cref{limitation} % \justificationTODO{}

\item {\bf Theory assumptions and proofs}
    \item[] Question: For each theoretical result, does the paper provide the full set of assumptions and a complete (and correct) proof?
    \item[] Answer:  \answerNA{} %\answerTODO{} % Replace by \answerYes{}, \answerNo{}, or \answerNA{}.
    \item[] Justification: This paper does not include theoretical results. %\justificationTODO{}

    \item {\bf Experimental result reproducibility}
    \item[] Question: Does the paper fully disclose all the information needed to reproduce the main experimental results of the paper to the extent that it affects the main claims and/or conclusions of the paper (regardless of whether the code and data are provided or not)?
    \item[] Answer: \answerYes{} % Replace by \answerYes{}, \answerNo{}, or \answerNA{}.
    \item[] Justification: We provide implementation details in Experiments section and \cref{app:experiments}.

\item {\bf Open access to data and code}
    \item[] Question: Does the paper provide open access to the data and code, with sufficient instructions to faithfully reproduce the main experimental results, as described in supplemental material?
    \item[] Answer: \answerYes{} % Replace by \answerYes{}, \answerNo{}, or \answerNA{}.
    \item[] Justification: We provide codes and data in the supplemental material.

\item {\bf Experimental setting/details}
    \item[] Question: Does the paper specify all the training and test details (e.g., data splits, hyperparameters, how they were chosen, type of optimizer, etc.) necessary to understand the results?
    \item[] Answer: \answerYes{} % Replace by \answerYes{}, \answerNo{}, or \answerNA{}.
    \item[] Justification: We provide details in both experiments and \cref{app:experiments}. %\justificationTODO{}

\item {\bf Experiment statistical significance}
    \item[] Question: Does the paper report error bars suitably and correctly defined or other appropriate information about the statistical significance of the experiments?
    \item[] Answer: \answerNo{} % Replace by \answerYes{}, \answerNo{}, or \answerNA{}.
    \item[] Justification: Error bars are not reported because it would be too computationally expensive for the LLM based method.%\justificationTODO{}

\item {\bf Experiments compute resources}
    \item[] Question: For each experiment, does the paper provide sufficient information on the computer resources (type of compute workers, memory, time of execution) needed to reproduce the experiments?
    \item[] Answer: \answerYes{} % Replace by \answerYes{}, \answerNo{}, or \answerNA{}.
    \item[] Justification: Implementation details are provided in \cref{app:experiments}. %\justificationTODO{}
    
\item {\bf Code of ethics}
    \item[] Question: Does the research conducted in the paper conform, in every respect, with the NeurIPS Code of Ethics \url{https://neurips.cc/public/EthicsGuidelines}?
    \item[] Answer: \answerNA{} % Replace by \answerYes{}, \answerNo{}, or \answerNA{}.
    \item[] Justification: We have not reviewed the NeurIPS Code of Ethics.%  \justificationTODO{}

\item {\bf Broader impacts}
    \item[] Question: Does the paper discuss both potential positive societal impacts and negative societal impacts of the work performed?
    \item[] Answer: \answerNA{} % Replace by \answerYes{}, \answerNo{}, or \answerNA{}.
    \item[] Justification: There is no societal impact of the work performed. %\justificationTODO{}
    
\item {\bf Safeguards}
    \item[] Question: Does the paper describe safeguards that have been put in place for responsible release of data or models that have a high risk for misuse (e.g., pretrained language models, image generators, or scraped datasets)?
    \item[] Answer: \answerNA{} % Replace by \answerYes{}, \answerNo{}, or \answerNA{}.
    \item[] Justification:  This paper poses no such risks. %\justificationTODO{}

\item {\bf Licenses for existing assets}
    \item[] Question: Are the creators or original owners of assets (e.g., code, data, models), used in the paper, properly credited and are the license and terms of use explicitly mentioned and properly respected?
    \item[] Answer: \answerNo{} % Replace by \answerYes{}, \answerNo{}, or \answerNA{}.
    \item[] Justification:  The base LLM is under \texttt{Llama 3.2 Community License Agreement}. The PLM is under \texttt{Apache license 2.0}. %\justificationTODO{}

\item {\bf New assets}
    \item[] Question: Are new assets introduced in the paper well documented and is the documentation provided alongside the assets?
    \item[] Answer: \answerNA{} % Replace by \answerYes{}, \answerNo{}, or \answerNA{}.
    \item[] Justification: The paper does not release new assets. %\justificationTODO{}

\item {\bf Crowdsourcing and research with human subjects}
    \item[] Question: For crowdsourcing experiments and research with human subjects, does the paper include the full text of instructions given to participants and screenshots, if applicable, as well as details about compensation (if any)? 
    \item[] Answer: \answerNA{} % Replace by \answerYes{}, \answerNo{}, or \answerNA{}.
    \item[] Justification: This paper does not involve crowdsourcing nor research with human subjects. %\justificationTODO{}

\item {\bf Institutional review board (IRB) approvals or equivalent for research with human subjects}
    \item[] Question: Does the paper describe potential risks incurred by study participants, whether such risks were disclosed to the subjects, and whether Institutional Review Board (IRB) approvals (or an equivalent approval/review based on the requirements of your country or institution) were obtained?
    \item[] Answer: \answerNA{} % Replace by \answerYes{}, \answerNo{}, or \answerNA{}.
    \item[] Justification: This paper does not involve crowdsourcing nor research with human subjects. %\justificationTODO{}

\item {\bf Declaration of LLM usage}
    \item[] Question: Does the paper describe the usage of LLMs if it is an important, original, or non-standard component of the core methods in this research? Note that if the LLM is used only for writing, editing, or formatting purposes and does not impact the core methodology, scientific rigorousness, or originality of the research, declaration is not required.
    %this research? 
    \item[] Answer: \answerNo{}{} % Replace by \answerYes{}, \answerNo{}, or \answerNA{}.
    \item[] Justification: The LLM is used only for writing and editing. %\justificationTODO{}

\end{enumerate}

\newpage
\appendix
\section{More details about node descriptions} \label{app:node_descriptions}
We provide more details about node descriptions in this section.
\subsubsection{TAGs}
In TAGs, we directly use their raw text as node descriptions. Since, ogbn-arxiv, PubMed and Cora are citation networks and every node represent a paper. We use title and abstract of every paper as their node feature. We provide several examples as below.

\begin{figure}[h]
  \centering
  \includegraphics[width=\textwidth]{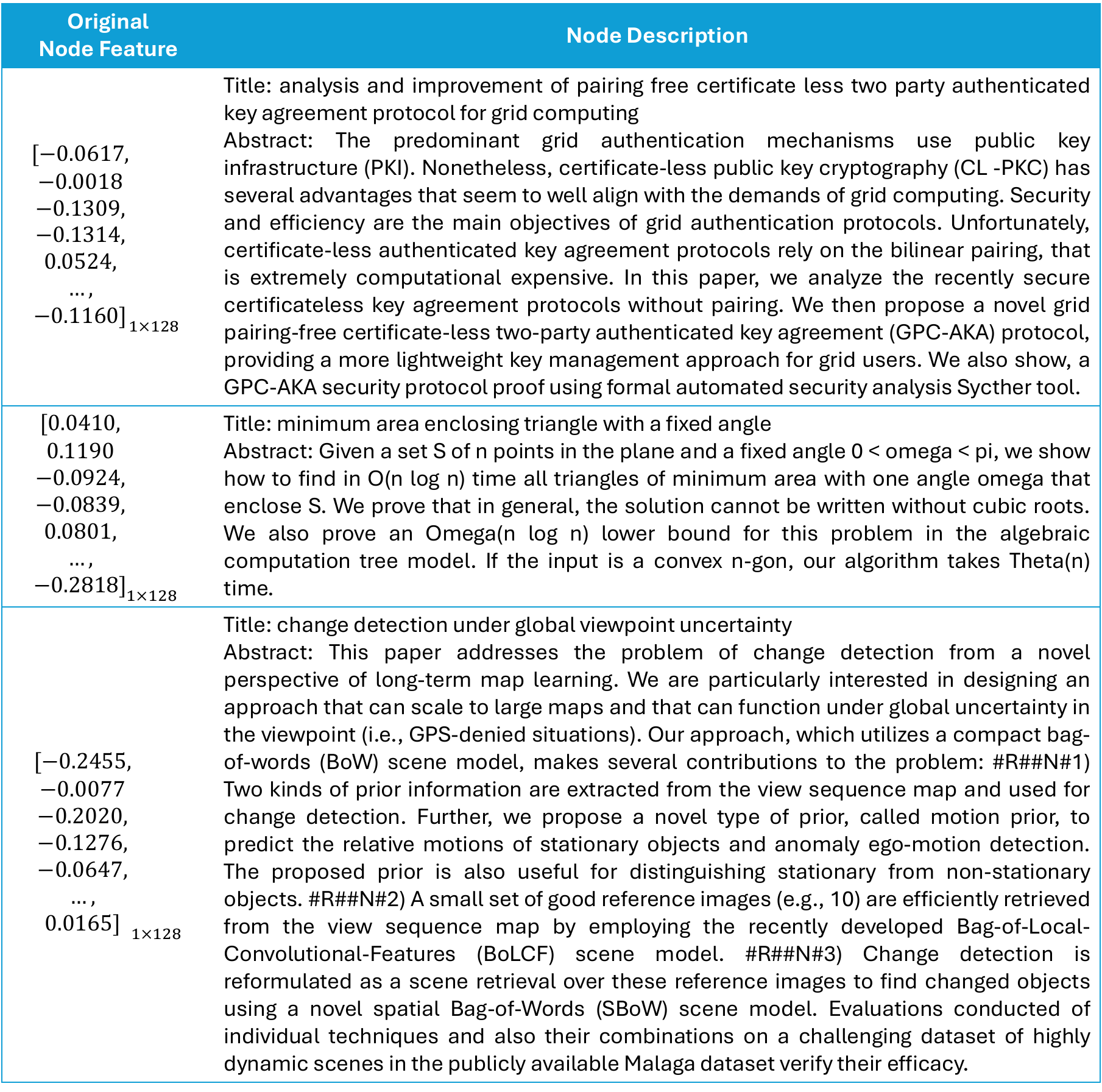}
  \caption{Three examples of the node description on ogbn-arxiv dataset.}
  \label{fig:arxiv_ns}
\end{figure}

\textbf{ogbn-arxiv}: We provide three examples of the node description on \cref{fig:arxiv_ns}.

\begin{figure}[h]
  \centering
  \includegraphics[width=\textwidth]{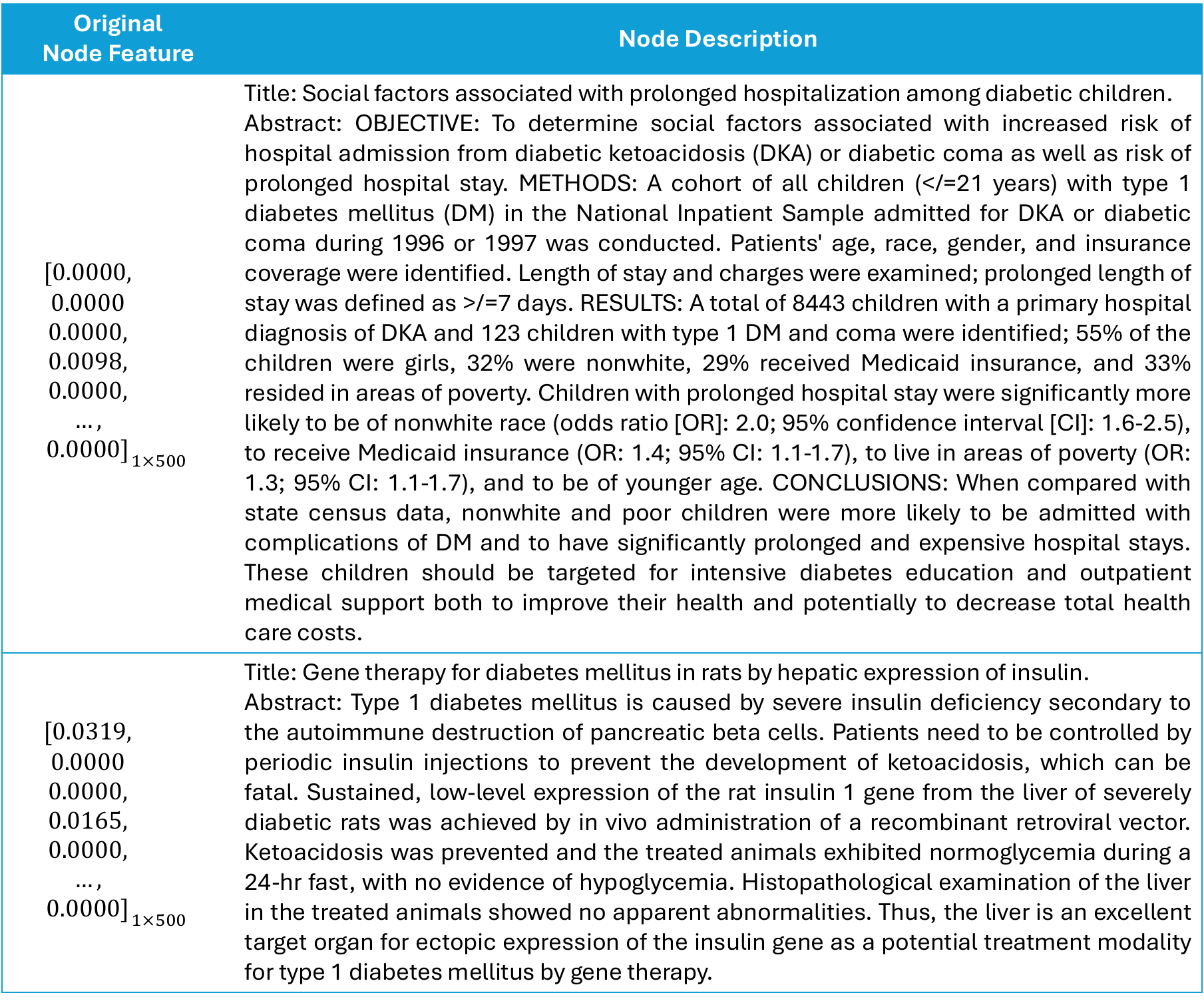}
  \caption{Two examples of the node description on PubMed dataset.}
  \label{fig:pubmed_ns}
\end{figure}

\textbf{PubMed}: We provide two examples of the node description on \cref{fig:pubmed_ns}.

\begin{figure}[h]
  \centering
  \includegraphics[width=\textwidth]{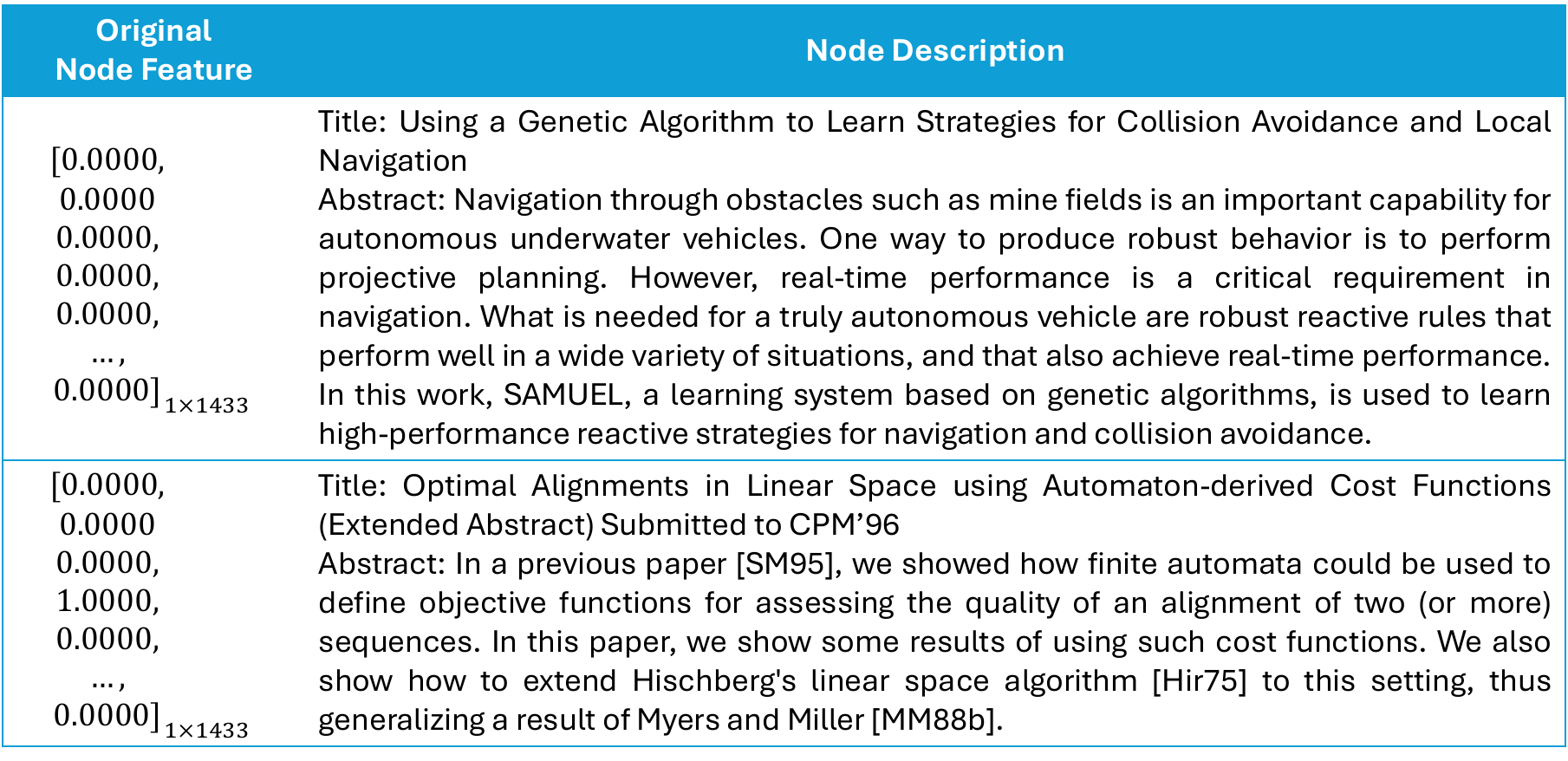}
  \caption{Two examples of the node description on Cora dataset.}
  \label{fig:cora_ns}
\end{figure}

\textbf{Cora}: We provide two examples of the node description on \cref{fig:cora_ns}.
\subsection{non-TAGs}
For non-TAGs, we need to generate descriptive language to characterize node features. Based on different original node feature, we provide different convert template as below.

\textbf{ogbg-molhiv}  is a molecular dataset in which each graph represents a molecule, and each node corresponds to an atom. Every node is associated with a 9-dimensional numerical feature vector. The semantic meaning of each dimension is detailed in \cref{tab:molhiv_meaning}. Based on the semantics of each dimension, we convert these numerical features into natural language using a structured template:
\begin{quote}
    This atom is [atomic name]. It has a [chirality type]. Its formal charge is [formal charge number]. The radical electrons of this atom is [number of radical electrons]. Its hybridization type is [hybridization type]. It connects [number of hydrogen atoms] hydrogen atoms. [This atom is part of an aromatic ring.] [This atom is part of a ring.] Its degree is [node degree].
\end{quote}

\begin{table}[h]
\centering
\caption{The meaning of every dimension on ogbg-molhiv dataset}
\label{tab:molhiv_meaning}
\begin{tabular}{@{}c|c@{}}
\toprule
\textbf{Index of Dimension} & \textbf{Meaning}                   \\ \midrule
1                           & Atomic number                      \\
2                           & Chirality type                     \\
3                           & Node degree                        \\
4                           & Formal charge                      \\
5                           & Number of connected hydrogen atoms \\
6                           & Number of radical electrons        \\
7                           & Hybridization type                 \\
8                           & Part of an aromatic ring?       \\
9                           & Part of a ring?                   \\
\bottomrule
\end{tabular}
\end{table}

\textbf{MUTAG} is also a molecular dataset. However, its node features are limited to the element type of each atom. In addition, the dataset provides edge type annotations that indicate whether a given bond is part of an aromatic ring. Based on this information, we construct natural language descriptions using the following template:
\begin{quote}
    This atom is [atomic name]. [This atom is part of an aromatic ring.] Its degree is [node degree].
\end{quote}

\section{More details about connector tuning} \label{app:connector_tuning}
The goal of node concept connector tuning is to effectively bridge the representations between the PLM and pretraind LLM.  We achieve this by training the LLM to reconstruct the original node description from its corresponding node concept embedding and to comprehend its high-level semantics. For each node description, we generate multiple single-turn conversation examples. 

\begin{figure}[h]
  \centering
  \includegraphics[width=\textwidth]{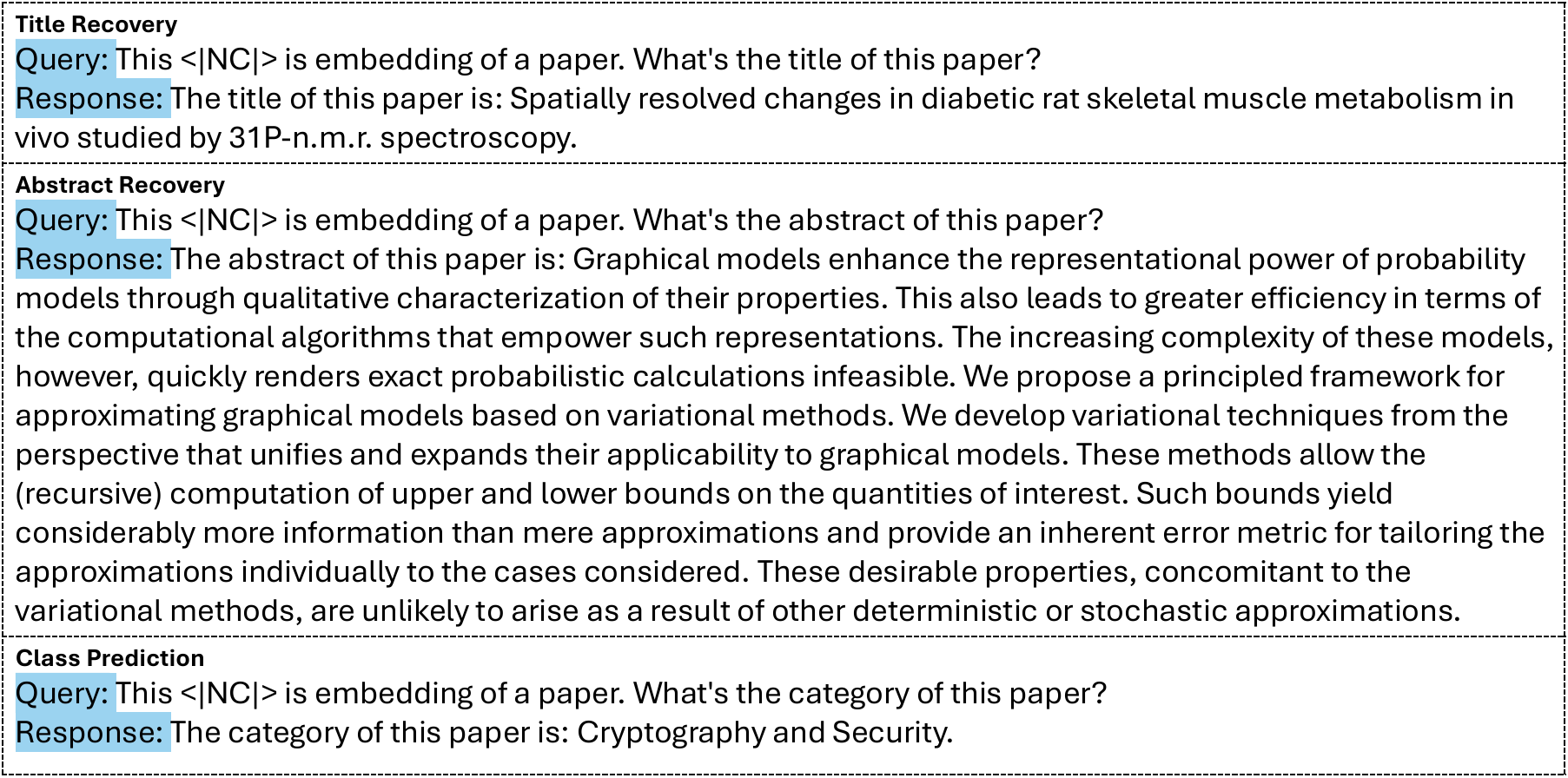}
  \caption{Examples of connector tuning for TAGs. To prevent data leakage, we ensure that the class prediction task is applied only to nodes from the training set.}
  \label{fig:connector_tuning_TAG}
\end{figure}

\textbf{TAGs} In TAGs, these conversations fall into three categories: 1) title recovery; 2) abstract recovery; 3) class prediction. Representative examples are provided in \cref{fig:connector_tuning_TAG}. To prevent data leakage, we ensure that the class prediction task is applied only to nodes from the training set. 

\begin{figure}[h]
  \centering
  \includegraphics[width=\textwidth]{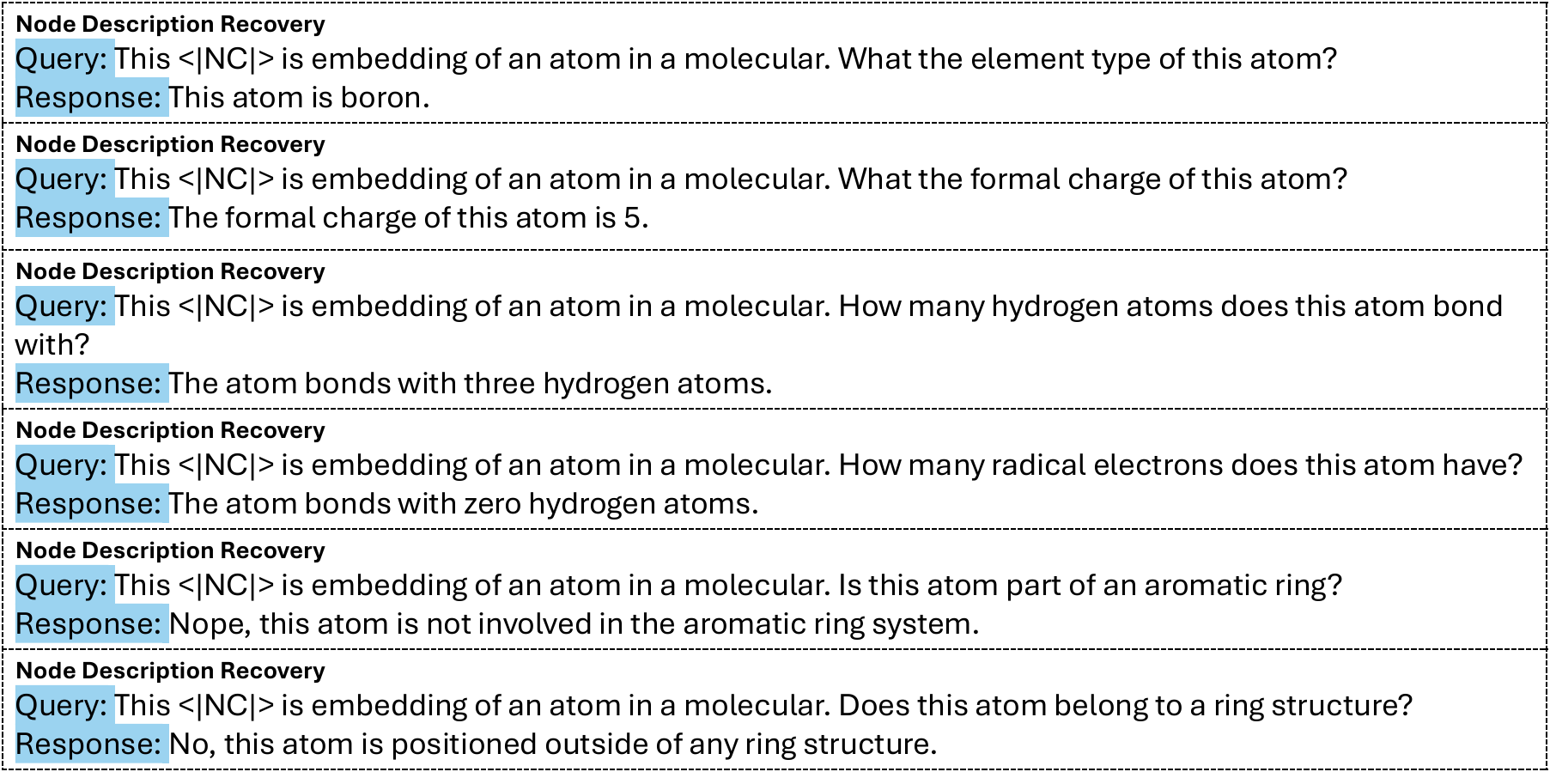}
  \caption{Examples of connector tuning for non-TAGs}
  \label{fig:connector_tuning_non_TAG}
\end{figure}

\textbf{Non-TAGs} For non-TAGs, we simply make LLM to reconstruct the original node description. We provide several example in \cref{fig:connector_tuning_non_TAG}.

\section{More details about reformulation graph tasks} \label{app:graph_tasks}
We reformulate all downstream graph tasks as human-readable queries. By casting these tasks as text-based comprehension problems, NOCL aligns naturally with the next-token prediction paradigm of LLMs, enabling them to generate task-specific outputs directly—without the need for specialized output heads or rigid task-specific architectures. This design not only accommodates standard classification tasks but also enables more expressive forms of graph reasoning, such as generating natural language explanations or answering free-form, graph-grounded questions. Below, we provide detailed templates and representative examples for standard classification tasks along with their corresponding expected responses.

\begin{figure}[h]
  \centering
  \includegraphics[width=\textwidth]{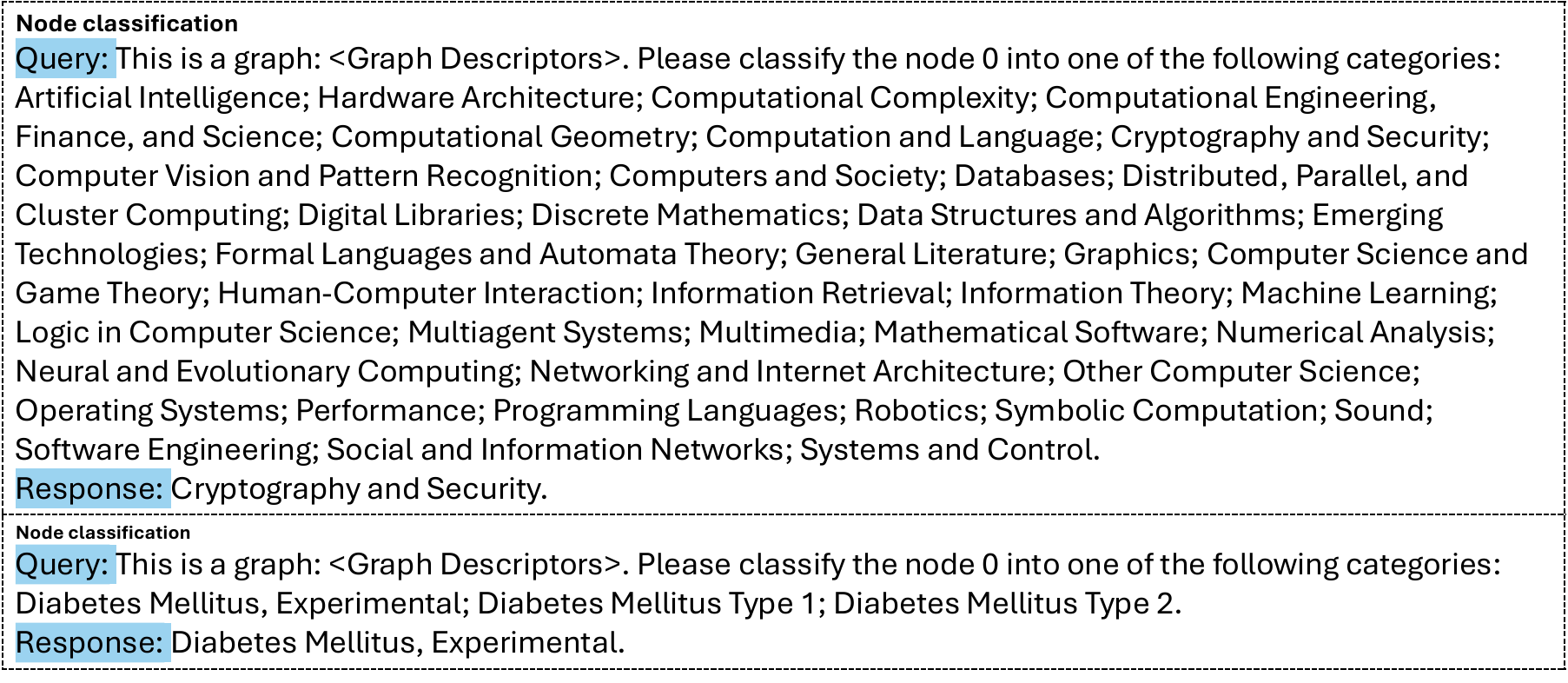}
  \caption{Examples of reformulation node classification task}
  \label{fig:node_examples}
\end{figure}

\textbf{Node classification} We directly prompt the LLM to predict the class of a target node from a predefined set of categories using the following template:
\begin{quote}
    Please classify the node [target node index] into one of the following categories: [predefined category 1, $\cdots$, predefined category n].
\end{quote}
We also provide two examples in \cref{fig:node_examples}.

\textbf{Link prediction} We prompt the LLM to determine whether two nodes should be connected using the following template:
\begin{quote}
    Should node [target node index 1] connect node [target node index 2]?
\end{quote}
The LLM is expected to respond with one of the following:
\begin{itemize}
    \item \textbf{"Yes, these two nodes should be connected."} -- indicating that an edge should exist between the two nodes.
    \item \textbf{"Nope, these two nodes have no relation."} -- indicating that no edge should exist.
\end{itemize}

\textbf{Graph classification} We prompt the LLM to assess whether a given graph exhibits a specific property. In ogbg-molhiv,the prompt is "Does the molecule have the ability to inhibit HIV virus replication?". For MUTAG, the question is "Is this molecule likely to exhibit mutagenic effects on Salmonella typhimurium?".

\section{Datasets} \label{app:datasets}
\begin{itemize}
    \item \textbf{ogbn-arxiv}: The ogbn-arxiv dataset comprises 169,343 Computer Science (CS) arXiv papers, each classified into one of the 40 categories. Each paper comes with a 128-dimensional feature vector obtained by averaging the embeddings of words in its title and abstract. The embeddings of individual words are computed by running the skip-gram model over the MAG corpus.
    \item \textbf{Cora}: The Cora dataset consists of 2708 scientific publications classified into one of seven classes. Each publication in the dataset is described by a 0/1-valued word vector indicating the absence/presence of the corresponding word from the dictionary. The dictionary consists of 1433 unique words.
    \item \textbf{PubMed}: The PubMed dataset comprises 19,717 scientific publications related to diabetes, each classified into one of three categories. The citation network includes 88651 links. Each publication is represented by a TF/IDF weighted word vector derived from a dictionary of 500 unique words.
    \item \textbf{ogbg-molhiv}: The ogbg-molhiv is a molecular property prediction dataset. It is adopted from the MoleculeNet. All the molecules are pre-processed using RDKit. Each graph represents a molecule, where nodes are atoms, and edges are chemical bonds. Input node features are 9-dimensional, containing atomic number and chirality, as well as other additional atom features such as formal charge and whether the atom is in the ring or not. The task is to predict the target molecular properties that whether a molecule inhibits HIV virus replication or not, as accurately as possible.
    \item \textbf{MUTAG}: The MUTAG dataset consists of 188 chemical compounds divided into two classes according to their mutagenic effect on a bacterium. Each graph represents a molecule, where nodes are atoms, and edges are chemical bonds. Input node features are 1 dimensional, indicating the element type of the atom.
\end{itemize}
A summary of the characteristics of the datasets is given in \cref{tab:ds_statistics}.

\begin{table}[]
\centering
\caption{Dataset statistics}
\label{tab:ds_statistics}
\resizebox{\textwidth}{!}{%
\begin{tabular}{@{}c|cccccccc@{}}
\toprule
Dataset     & Task                 & \# Graph & \# Node & \# Ave. Node & \# Edge & \#Ave Edge & \# Feature & \# Class \\ \midrule
ogbn-arxiv  & node classification  & 1        & 169343  & -            & 2484941 & -          & 128        & 40       \\ \midrule
Cora &
  \begin{tabular}[c]{@{}c@{}}node classification\\ /\\ link prediction\end{tabular} &
  1 &
  2708 &
  - &
  13264 &
  - &
  1433 &
  \begin{tabular}[c]{@{}c@{}}7\\ /\\ 2\end{tabular} \\ \midrule
PubMed &
  \begin{tabular}[c]{@{}c@{}}node classification\\ /\\ link prediction\end{tabular} &
  1 &
  19717 &
  - &
  88651 &
  - &
  500 &
  \begin{tabular}[c]{@{}c@{}}3\\ /\\ 2\end{tabular} \\ \midrule
ogbg-molhiv & graph classification & 41127    & 1049163 & 25.51        & 2259376 & 54.94      & 9          & 2        \\ \midrule
MUTAG       & graph classification & 188      & 3371    & 17.93        & 7442    & 39.59      & 1          & 2        \\ \bottomrule
\end{tabular}%
}
\end{table}
\section{More details about experiments} \label{app:experiments}
To effectively reproduce our experiments, we provide the code at \url{https://anonymous.4open.science/r/NodeConceptLLM-1B7E}
% \section{Technical Appendices and Supplementary Material}
% Technical appendices with additional results, figures, graphs and proofs may be submitted with the paper submission before the full submission deadline (see above), or as a separate PDF in the ZIP file below before the supplementary material deadline. There is no page limit for the technical appendices.

\end{document}